\documentclass[conference]{IEEEtran}
\usepackage{fancyhdr}
\usepackage{algorithm}
\usepackage{algorithmic}

\usepackage{graphics} 
\usepackage{float}
\usepackage{epsfig}

\usepackage{amsmath} 
\usepackage{amssymb,amsfonts}

\usepackage{mcite}

\ifCLASSINFOpdf
\else
\fi
\begin{document}
%
\title{Monocular Urban Localization using Street View }


\author{Li Yu$^{*,\dagger}\hspace{5mm}$Cyril Joly$^{*}$ \hspace{5mm}Guillaume Bresson$^{\dagger}\hspace{5mm}$ Fabien Moutarde$^{*}$ 
\\$^{*}$Centre of Robotics, MINES ParisTech, PSL\\
60 Blvd St Michel, 75006, Paris, France\\
Email: firstname.name@mines-paristech.fr\\$^{\dagger}$Institut VEDECOM\\
77 Rue des Chantiers, 78000, Versailles, France\\
Email: firstname.name@vedecom.fr}

\maketitle

\thispagestyle{fancy}
\fancyhead{}
\lhead{}
\lfoot{}
\cfoot{}
\rfoot{}
\renewcommand{\headrulewidth}{0pt}
\renewcommand{\footrulewidth}{0pt}

\begin{abstract}
This paper presents a metric global localization in the urban environment only with a monocular camera and the Google Street View database. We fully leverage the abundant sources from the Street View and benefits from its topo-metric structure to build a coarse-to-fine positioning, namely a topological place recognition process and then a metric pose estimation by local bundle adjustment. Our method is tested on a $3$ km urban environment and demonstrates both sub-meter accuracy and robustness to viewpoint changes, illumination and occlusion. To our knowledge, this is the first work that studies the global urban localization simply with a single camera and Street View.
\end{abstract}

\IEEEpeerreviewmaketitle

\section{Introduction}
The past decade has seen a substantial progress in the autonomous vehicles field. However, the metric localization in urban environment is still one of crucial challenges \cite{fuentes2015visual}. The Global Positioning System (GPS) suffers from a precision degradation and even fails in ``urban canyons''.  The Simultaneous Localization and Mapping (SLAM) requires to build a large consistent map before the localization. The data fusion technique based on GPS, proprioceptive sensors and existing maps, is still tricky to evaluate the confidence of different data sources and its localization quality depends a lot on sensors' price. 
Alternatively, Geographical Information Systems (GIS), such as Google Street View, Mappy, Navteq, etc., are more and more precise and offer a unified global representation of our world with visual, topological, spatial and geographic information \cite{zhang2006image}. This motivates us to fully take advantage of the abundant sources in GIS to realize a metric localization in urban contexts. In this paper, Google Street View is adopted for its world-wide coverage, public accessibility, high resolution geo-referenced panoramas and well-calibrated depth \cite{anguelov2010google}. We develop a coarse-to-fine position estimation system. At the coarse level, a topological localization is addressed by a place recognition algorithm. Namely, a query image captured by our camera equipped on vehicle can be associated with geotagged Street View images when sharing high appearance similarities. The vehicle's position is thus restricted on an intersected area around the geodetic position of corresponding Street View cameras. At this stage, an efficient retrieval method is described to shrink the search time. Then, since the depth maps are integrated in Street View, we can refine the before-mentioned location by computing the vehicle's 6 degrees of freedom (DoF) transformation \textit{w.r.t} Street View images by solving the Perspective-n-Point (PnP) problem \cite{bolles1981ransac}. Finally, a local Bundle Adjustment (LBA) \cite{mouragnon2006real} is applied to improve the consistency of the localization regarding to the multiple Street View images retrieved. 

In fact, extensive efforts have been made to solve the urban localization by using GIS data such as available maps \cite{levinson2007map}, street network layout \cite{floros2013openstreetslam}, geotagged traffic signs \cite{qu2015vehicle}, satellite images \cite{dogruer2008global} and 3D texture city models \cite{majdik2015air}. They focus on one of the above sources as a single strong constraint to optimize their tasks and rarely consider the entire topo-metric information in the GIS, for instance, Street View is commonly employed only as an image database input to solve the place recognition problem \cite{chen2011city}. Numerous papers pay attention to the localization in a large-scale environments,  with the view point change or under cross-season or light variation \cite{vaca2012city,torii201524}. Their pipeline is to construct a Bag-of-Words (BoW) model \cite{sivic2003video}, query an input image within the model and localize the place by comparing images vectors' similarity. On the other hand, in our approach, we propose to consider the place recognition problem as a primary selection to feed the metric localization. Moreover, our approach consider the multiple database constraints to speed up the searching instead of a pure statistic BoW matching. 

In a similar fashion to our work, Agarwal et al. \cite{agarwal2015metric} realize an urban localization with a sub-meter accuracy by modeling a two-phase non-linear least squares estimation. They first recover 3D points position from a mono-camera sequence via an optical flow and then compute the rigid body transformation between the Street Views and the estimated points. Street Views are learnt as a BoW model and retrieve efficiency is improved via an inaccurate GPS input to narrow the searching radius. Their key contribution is to localize globally on Google Maps without a heavy map construction. This method still follows a SLAM manner. A map building is omitted but numerous 3D scene points must be estimated and registered as usual. The Street View imagery serves as keyframes to solve the loop closing problem. Rather, our method fully depends on the Street View and the only input is just a stream of mono-camera images without odometry or GPS information. 

\section{Method}
In this section, we describe the coarse-to-fine localization system in detail. A pseudo-code is provided in Algorithm \ref{overALg} to give an overview. Note that the system is divided into two phases. The Street View panoramas are prepared and their rectilinear views are generated offline (Line $1$ to $7$) prior to an actual online vehicle localization (Line $8$ to $18$). In the offline part, all generated imagery is trained as a BoW database. A symmetric comparison matrix is computed for the database so as to reduce the time to retrieve close images in online phase. It also serves as a index reference for the following optimization steps. In the online stage, for every vehicle query image, we extract top similar images from the database as the topological localization and then estimate relative poses between each other. Finally, a LBA is employed with all estimated poses and corresponding 3D-to-2D matching constraints to obtain the global metric localization.
\begin{algorithm}
\caption{Metric global localization in the urban area }
\label{overALg}
\begin{algorithmic}[1]
\renewcommand{\algorithmicrequire}{\textbf{Input:}}
\renewcommand{\algorithmicensure}{\textbf{Output:}}
\renewcommand{\algorithmiccomment}[1]{\hskip2em$\triangleright$ #1}
\REQUIRE Street View panoramas  $\textbf{S}=\left\lbrace S_{1}, S_{2},\dots, S_{n}\right\rbrace$ and \\ \enskip \enskip \enskip their depth maps $\textbf{D}=\left\lbrace D_{1}, D_{2},\dots, D_{n}\right\rbrace$ 
\REQUIRE Query images $I_{q} =\left\lbrace I_{1}, I_{2},\dots, I_{m}\right\rbrace$ captured by a \\ \enskip \enskip \enskip vehicle driving in the city
\ENSURE Global location of the vehicle
\FOR{$i \gets 1$ \TO $n$}
\STATE Rectilinear processing of $\textbf{S}$ and $\textbf{D}$ 
\STATE Feature extraction and BoW training 
\STATE Construction of $DB$ as the final BoW dictionary
\ENDFOR
\STATE Calculation of the intra-distance matrix $D_{DB \times DB}$ and speed-up matrix $D_{DB \times r}$  \COMMENT{\textit{cf.} Section \ref{dbdb}}
\STATE \textbf{\textit{Up to here the algorithm is implemented offline.}}
\FOR{$t \gets 1$ \TO $n$}
\STATE Parametrize $I_{q}$ by BoW
\STATE Search $I_{q}$'s the best similar image $I_{r}$ in $DB$ 
\STATE Get the top $k$ similar database images of $I_{r}$ in $D_{DB \times r}$
\STATE Histogram Equalization: $I_{t/r}^{'}=E\left(I_{t/r}\right)$ 
\FOR{$j \gets 1$ \TO $k$}
\STATE{Feature matching}
\STATE{Pose estimation $\Theta_{j}$}  \COMMENT{\textit{cf.} Equation \ref{localO}}
\ENDFOR
\STATE{$\Theta^{\star}= LBA\left\lbrace\Theta_{j}\right\rbrace$}
\STATE{Recover the global position at step $t$}
\ENDFOR

\end{algorithmic}
\end{algorithm}
\subsection{Data preparation and generation}
\label{datapre}
\begin{figure*}
	\centering
	\includegraphics[width = \textwidth]{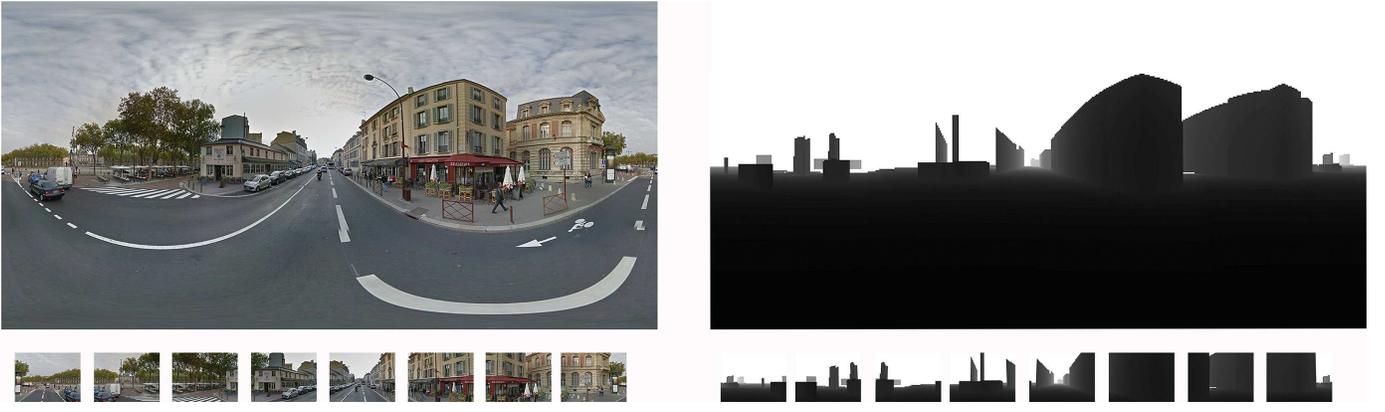}
	\caption{An example of Street View panorama (top-left) and its associated depth map (top-right)  in Versailles, France. The below $8$ rectilinear images and depth maps are extracted from the above ones by the back projection. The $8$ virtual pinhole cameras, are configured similarly to the vehicle's camera, including the same focal length and image size. It creates overlapping views. }
	\label{pano}
\end{figure*}

\begin{figure}
	\centering
	\includegraphics[scale = 0.5]{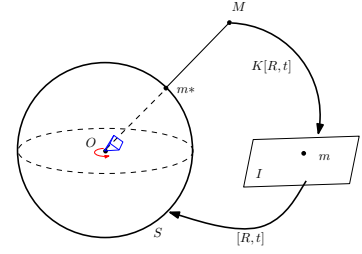}
	\caption{Back-projection model: 8 virtual cameras are constructed at point $\textbf{O}$ and pixels in the image plane $\textbf{I}$ are bilinearly interpolated from panorama sphere. The yaw offset changes according to the red arrow direction. }
	\label{backproj}
\end{figure}

Google Street View is a rich street-level GIS with millions of panoramic imagery and depth maps captured all over the world (see Figure \ref{pano}) \cite{anguelov2010google}. Every panorama, geotagged with an accurate GPS position, is registered a $13,312\times6,656$ pixel resolution by capturing a $360^{\circ}$ horizontal and $180^{\circ}$ vertical field-of-view (FoV). It is composed and projected by rectilinear images from several cameras. Generally, there is a panorama every $6$ to $15$m with a nearly uniform street coverage. The associated depth map stores the distance and orientation of various points in the scene via laser range scans or using image motion methods (optical flow). It only encodes the scene's dominant surfaces by its normal direction and its distance, allowing to map building fa\c{c}ades and roads while ignoring smaller entities such as vehicles or pedestrians. For the sake of bandwidth saving, it is sampled down to $512\times256$ pixels but recovering the similar size to the panorama is easy. The GPS position of Street Views is highly precise due to a careful global optimization, while the depth map provides a 3D structure of the scene with a relatively low accuracy \cite{anguelov2010google}. Depending on the route planner built in Street View, all panoramic images and depth maps can be downloaded iteratively.
The panoramas ought to be transformed into a set of overlapping or unrelated cutouts to reduce the large angle distortion. Google offers a public API \cite{googleapi} to extract rectilinear images by setting a virtual camera's parameters, including FoV, pitch and yaw. Currently, most of relative researches rely on this API to generate perspective images. Instead, we build a back-projection model to realize a more robust and flexible extraction, as depicted in Figure \ref{backproj}. We assume $8$ virtual pinhole cameras with the camera matrix $K$ are mounted in the centre of a unit sphere $\textbf{S}$ with a user fixed pitch direction and the following yaw directions $\left[0^{\circ}, 45^{\circ}, ..., 360^{\circ}\right]$. The number of virtual cameras, intrinsic matrix and pitch/yaw degree are free to select, yet empirically the more identical they are to the actual on-board camera, the better performance expected. That is one of the reasons why we develop our own rectilinear model.
Consider a 3D point $\textbf{M}\in\mathbb{R}^{3}$ in homogeneous coordinates using spherical parametrization $(\theta, \phi, \rho)$, denoted as:
\begin{equation}
\textbf{M}=\begin{bmatrix}
\rho\cos(\theta)\sin(\phi)\\
\rho\sin(\theta)\sin(\phi)\\
\rho\cos(\phi)\\
1
\end{bmatrix}
\end{equation}
Its projection on the unit sphere is represent by $\textbf{m}^{*}$ 
\begin{equation}
\textbf{m}^{*}=\frac{\textbf{M}}{\lVert\textbf{M}\rVert}
=\begin{bmatrix}
\cos(\theta)\sin(\phi)\\
\sin(\theta)\sin(\phi)\\
\cos(\phi)\\
1
\end{bmatrix}
\end{equation}
and its projection on a virtual camera image plan by $\textbf{m}$.
\begin{equation}
\textbf{m}=K\left[\textbf{R},\textbf{t}\right]\textbf{M}
\end{equation}
where the camera extrinsic matrix $\left[\textbf{R},\textbf{t}\right]$ is deduced from the above configuration. The translation $\textbf{t}$ equals zero when camera is fixed in the same point. The perspective virtual views can be synthesized by the rays tracing back to the panorama with a bilinear interpolation around the back-projected points. The warp process is noted as $\omega$.
\begin{equation}
\textbf{m} \stackrel{\omega}{\longrightarrow}\textbf{m}^{*}
\end{equation}
The corresponding perspective depth map can be addressed likewise. A generation example from a Street View panorama of Versailles in France is illustrated in Figure \ref{pano}. 

In the experiment, we have totally collected a test set of $3655$ query images captured by a MiPSee camera on the vehicle with a $57.6^{\circ}$ FoV. Images were captured at the city centre of Versailles. The ground truth at each localization was recorded by a centimeter-level real-time kinematic GPS (RTK-GPS).

\subsection{Database construction and retrieval acceleration}
\label{dbdb}
After the artificial generation, the Street View database becomes 8 times larger in quantity. We use BoW retrieval techniques to represent database images as numerical vectors quantized by feature descriptors, and to perform a hierarchical clustering (K-means) of the image descriptors in a tree structure. After the weighing strategies, the whole tree is referred to as a dictionary and its leaves as visual words. A topological localization is estimated according to a distance criterion based on the vectors similarity. In our test, we use the Term Frequency-Inverse Document Frequency (TF-IDF) reweighing and the efficient cosine similarity distance metric. In particular, considering the dynamic changes in viewpoints, illumination and occlusions, we construct two independent dictionaries generated from the SIFT and MSER detectors, and then normalize them. The final dictionary can take into account both local and regional feature descriptors.  

\begin{figure}
	\centering
	\includegraphics[width = \linewidth]{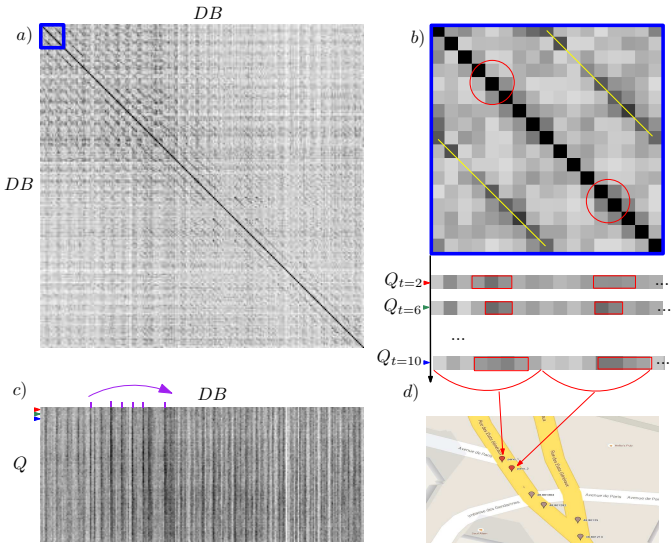}
	\caption{Illustration of distance matrices and intra-relations: $a$) The intra-database symmetric matrix $D_{DB\times DB}$. The rows and columns represent $240$ rectilinear views generated from $30$ panoramas in a $350$m urban route, and the matrix intensity is computed by mapping the cosine similarity from $\left[0,1\right]$ to $\left[0,255\right]$, therein, the darker pixels depict the higher similarities and diagonal values are always equal to $255$. $b$) The above matrix is a close-up view \textit{w.r.t} the blue rectangle in $a$), it shows two obvious yellow lines with high intensity, that are parallel to the diagonal line indicating a location regularity for similar images in the matrix. $c$) The distance matrix $D_{DB\times Q}$ rows represent the $240$ database images and columns represent a query sequence with $100$ frames. Several darkest vertical lines are highlighted by purple marks, meaning that a short query sequence can find its most similar database images only in few range of the database as shown by the purple arrow. Considering the candidate location at query time steps $t=2, 6, 10$, we compare their close-ups to the database shown in $b$) and conclude that even in the same panorama, the similar candidates sparsely focus on some certain yaw offsets. The red cycles and rectangles represent top similar candidates from the same panorama. $d$) The panoramas are searched at $3$ time steps in $b$). }
	\label{symMat}
\end{figure}

As a rule of thumb, the bigger the database is, the slower will be the information retrieval from it. The aim now is to facilitate the run-time search even if a metropolis database is constructed. In a natural manner, we explore the intra database similarities and integrate the potential topological information to reduce the computational cost. Figure \ref{symMat} shows the similarity between images within the whole database $D_{DB\times DB}$ as well as the relationship between the database and query images $D_{DB\times Q}$. The results, detailed in the figure caption, show a certain regularity to locate top similar images \textit{w.r.t} a query image. To some degree, these regularities reflect the topology and inner connections in the GIS, e.g. all database images are downloaded successively according to the Street View route planner. Intuitively, neighboring panoramas share similar appearances and correspondingly their similar rectilinear images ought to situate in an analogous yaw or auxiliary angle of yaw offset (only for two-way streets). That is the reason why many yellow parallel lines and vertical dark lines lie in $D_{DB\times DB}$ and $D_{DB\times Q}$ respectively. 
All these regularities make our database a sparse ``searching map", which are advantageous to significantly reduce the number of image comparisons in the online phase. It is summarized as follows:
\begin{itemize}
\item In a piecewise route database $D_{DB\times DB}$, for each database image $I_{r}$ we register its top $k$ similar database images' $I_{0\dots k}$ location in an array  $D_{DB\times r}$.
\item A maximum appearance distance $dist_{max}$ is defined as the farthest radius around a panorama where its neighboring panoramas share an overlapping view. Normally, $6$ to $8$ nearest panoramas share a common scene around $50$ to $70$ meters. In $D_{DB\times DB}$, the $dist_{max}$ can be fixed as $48$ or $64$ matrix index steps.
\item With the help of the route planner, we can localize the first incoming query image easily by searching the database image around the starting point. Then for an ordinary query image, its searching range can be narrowed to all $D_{DB\times r}$ within the threshold $dist_{max}$.
\end{itemize}
According to our experiment, the above approach can avoid nearly $90\%$ of the comparisons without any retrieval loss compared to a whole database search. 

\subsection{Global metric localization}

In the previous sections, a candidate set of images has been retrieved from database. In order to realize a metric localization, high-inlier feature correspondence between camera images and the candidate set should be guaranteed. Other than a statistic comparison with the BoW, the conventional perspective matching is done by following this kind of pipeline \cite{hartley2003multiple}: $\left(a\right)$ a set of keypoints are extracted in both query and database images, e.g. SIFT; $\left(b\right)$ their descriptors are matched by nearest neighbor search algorithms, e.g. FLANN; $\left(c\right)$ matching outliers are rejected by the ratio test and geometric verification using constraints from the homograhy or the fundamental matrix, e.g. RANSAC $8$-point algorithm. Unfortunately, we found this pipeline is unstable and inaccurate because of important changes from viewpoint, illumination and occlusion in urban area. 

To counter this effect, we use the histogram equalization \cite{hummel1975histogram} to enhance the contrast of all images prior to feature extraction. Also the Virtual Line Descriptor (kVLD) \cite{liu2012virtual} is applied to determine the inlier feature point correspondences. It is a SIFT-like descriptor by signing virtual lines to the points with geometrical consistency.  The algorithm computes and matches a $k$ connected virtual line graph to reject the outliers, see Figure \ref{match}.

\begin{figure}
	\centering
	\includegraphics[scale =0.4]{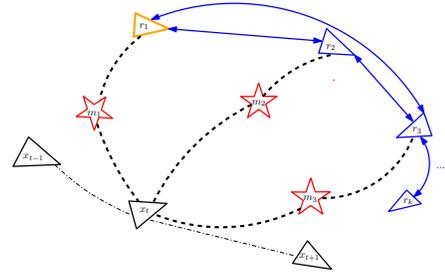}
	\caption{Illustration of Local Bundle Adjustment to estimate the global position of the vehicle. The vehicle, the best similar database image and other top $k-1$ similar database images are the triangles respectively colored in black, orange and blue. The red stars represent good matching features between the query and database images. }
	\label{localO}
\end{figure}

Figure \ref{localO} depicts our global metric localization process. In this system, the vehicle captures images at consecutive camera states $x_{t-1}$ and $x_{t}$. No odometric input is integrated between successive states. At the state $x_{t}$, the best database image $r_{1}$ associated to the query image is retrieved through the topological localization. As mentioned before, via the database $D_{DB\times r}$,  the top $k-1$ analogous database images to the best one are found and denoted as $\left[r_{2}, r_{3}, \dots,r_{k}\right]$. Naturally, we suppose that the query image shares the overlapping view with these $k-1$ images as well. The constraints between them are found thanks to the accurate matching features. With the help of depth maps in the database, the 6 DoF pose of the vehicle $\Theta=\left(\textbf{R},\textbf{t}\right)$, parametrized in Lie algebra $\textbf{SE}\left(3\right)$, can be computed by minimizing the reprojection error between a pair of images, i.e.:

\begin{equation}
\Theta^{\star}=\arg\min_{\Theta}\sum_{i}\pi\left(\lVert\textbf{m}_{i}-\textbf{P}\left(\textbf{M}_{i},\Theta\right)\rVert\right)
\end{equation}

where $\textbf{P}\left(\textbf{M}_{i},\Theta\right)$ is the image projection from the scene point $\textbf{M}_{i}$. The 3D-to-2D correspondence is improved by RANSAC and $\pi$, as a M-estimator based on Tukey Biweight function \cite{zhang1997parameter}, is used to improve convergence and to cope with wrong correspondence.
\begin{equation}
\label{rho}
\pi\left(x\right)=
\begin{cases}
t^{2} \mathbin{/}6\left(1-\left[1-\left(\frac{x}{t}\right)^{2}\right]^{3}\right) &\mbox{if $|x|\leq t$  }\\
t^{2} \mathbin{/}6 &\mbox{if $|x|>t$ }
\end{cases}         
\end{equation}
Finally, $k$ poses and corresponding image pairs are put into a LBA to refine the vehicle's pose and its global position.

We use the optimization framework $g2o$ \cite{kummerle2011g} as our non-linear least squares solver. The error regarding to the feature detection is assumed to follow a Gaussian distribution. The $k$ geotagged views extracted from panoramas are constrained in Figure \ref{localO} and their camera configuration are given. We can use them to verify the performance of our metric localization since their absolute positions are known. Moreover, the parameters of the LBA, can be first initialized by this verification as well.

\section{Experimental Evaluation and Discussion }
In order to evaluate the performance of our algorithm, we test on several city streets and record query images when the RTK-GPS reaches a localization within $20$cm precision. Using this kind of RTK-GPS positions as the ground truth is sufficient for our metric accuracy qualification. In the experiment, our vehicle is mainly tested on the roads around the area with GPS coordinates $\left[48.801631,2.131509\right]$. The intrinsic parameters of Google virtual cameras are fixed according to our own MiPSee camera. Only one-side of city fa\c{c}ades are captured.

\begin{figure}
	\centering
	\includegraphics[scale =0.65]{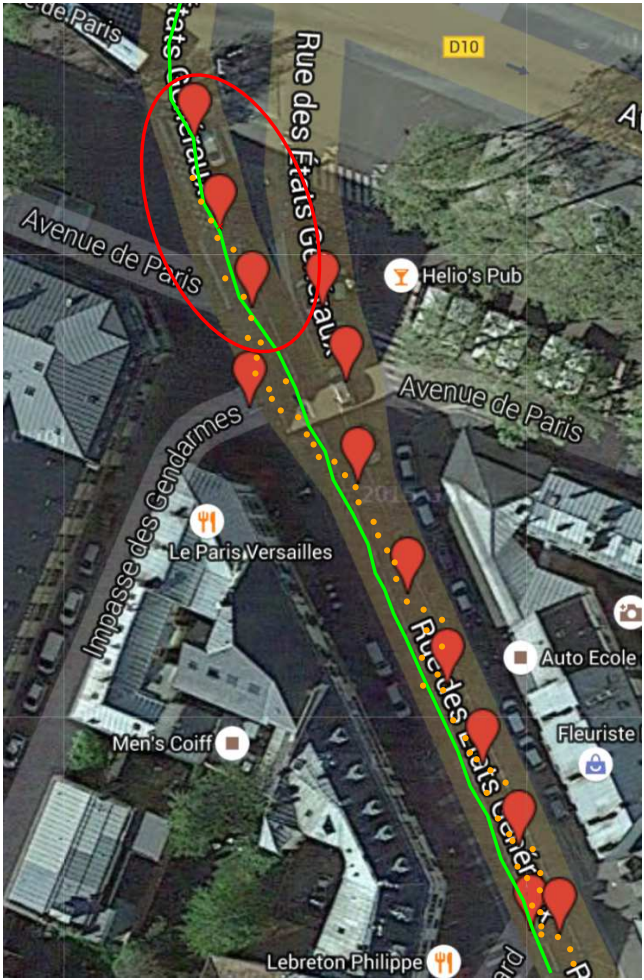}
	\caption{Bird's-eye view of the metric global localization. The red Google pointers represent the locations of the Street View cameras. The orange dots mark the estimated positions of the monocular camera. The green line illustrates RTK-GPS ground truth and  the red circle shows the topological bound around a panorama denoted by the cross mark.}
	\label{meloc}
\end{figure}

The performance of our coarse-to-fine algorithm depends closely on both topological and metric localization part. In fact, if the coarse topological localization fails, the further metric part based on it works in vain. However, in city scenario, it is difficult to construct a strict ground truth to evaluate the topological localization. In \cite{majdik2015air} and \cite{majdik2013mav}, authors proposed a manual labeling way to build up a confusion matrix as the ground truth. This method is too time consuming to employ as our test dataset contains more than 3000 frames. As expected, the visual overlap must exist between the query and database image if the BoW algorithm works well. Therefore we can evaluate the topological localization based on the geometric consistency. If the inlier-match number between two images is larger than $12$, we can get a bound with a $19.8$m radius around the Street View in which the corresponding query images lie, as shown by the red circle in Figure \ref{meloc}. The bound radius distance is calculated between the RTK-GPS of query images and the geo-data of retrieved Street View. After manually verifying the pairs matches along a street with $33$ panoramic Street Views, our method is able to recognize $100\%$ visual similar database images corresponding to query images. The $264$ rectilinear Street Views are trained to a mixed 5000 SIFT and 2000 MSER visual words.

Based on the previous topological localization, we choose the EPnP-RANSAC to compute the pose between each query monocular image and matched Street View. The solution is selected with a highest consensus by setting a minimum reprojection error within $3$ pixels and maximum inliers. Then a $g2o$ framework is used to optimize all computed solutions. This process is realized under the cartesian coordinates with the Lambert conformal conic projection and the result is converted back to geo-coordinates under the WSG84.

\begin{figure}
	\centering
	\includegraphics[scale =0.43]{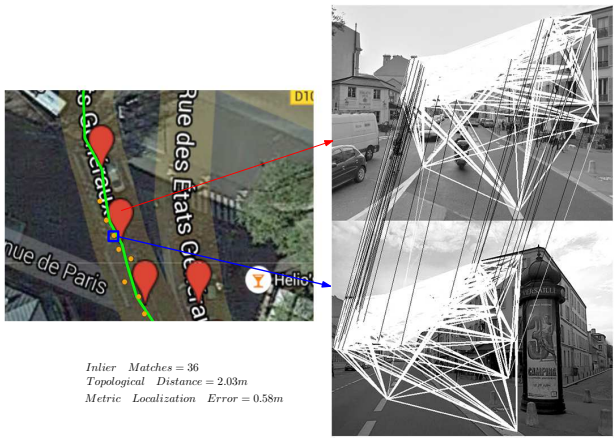}
	\caption{An example of matches between a monocular image and its retrieved Street View image by kVLAD descriptors. The metric localization reaches a $0.58$m accuracy by refining from the topological localization ($2.03$m).}
	\label{match}
\end{figure}

Figure \ref{meloc} shows the metric global localization in a $287m$ city street where $423$ query images and 13 panoramic Street Views are used. There are $58$ metric positions obtained within an error of $6.5$m \textit{w.r.t} the RTK-GPS and $58.6\%$ keeps within a $2$m accuracy. An example of matches between the query and retrieved Street View is depicted in Figure \ref{match}.

\begin{figure}
	\centering
	\includegraphics[scale =0.35]{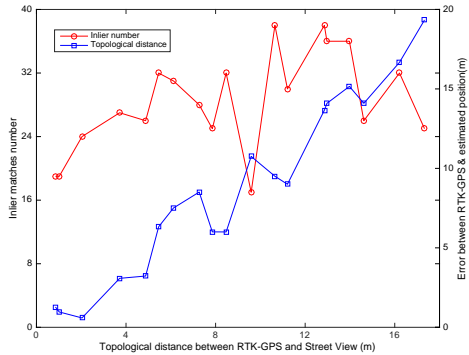}
	\caption{Metric error analysis \textit{w.r.t} inlier-match numbers and the topological distance.}
	\label{result}
\end{figure}

Although the accuracy is higher than the regular autonomous GPS (average error of $8$m ) \cite{wing2005consumer},  readers may still wonder the discontinuity (only $58/423$ frames localized) and drifts.  The cause of errors during the metric localization thus is analyzed in two aspects:  the topological distance between the chosen Street View and RTK-GPS ground truth, and the inner matches between the query and Street View image. The inlier-match number reflects the feature quantity that is used in the 3D-to-2D pose estimation process, and the real topological distance between the query and Street View image can describe the feature quality. Commonly, the less or the further the features are tracked, the less accurate the localization is. We select all query images that retrieve the same Street View located at the coordinates $\left[48.801631,2.131509\right]$ and obtain the relationships in Figure \ref{result}.  Our effective BoW algorithm guarantees the visual overlaps between query images and retrieved Street Views around $20$m bound. In this way, the inlier-match number affects little to the drifts. However, the errors appear in a cumulative trend when topological distance increases between query and Street View. Despite the optimization improves the accuracy, too many further features only show small motions and still cause disastrous errors even passing the bound of topological localization. As a result, we eliminate lots of localization like this and cause unavoidable localization discontinuity. Naturally the synthesis view from the Street View can be created to augment nearer features and enhance the accuracy. This will be one of our future work.

\section{Conclusion}
To conclude, this paper presents a metric urban localization by matching 3D Street View points to a monocular camera. Our approach requires neither the construction of a consistent map nor the prior visit of the environment.  This simple set-up makes our algorithm affordable and easy to be widely deployed in real urban localization. In order to realize a reliable localization, we throughly explores multiple information of Street View and strictly respects their intra relationships. Our technique demonstrates both the high accuracy \textit{w.r.t} the Street View and robustness in complicated urban environments.

\section*{Acknowledgment}
This work is supported by the Institut VEDECOM in France under the autonomous vehicle project and the China Scholarship Council.

\bibliographystyle{IEEEtran}
\bibliography{icarcv}

\end{document}